\documentclass[letterpaper, 10 pt, conference]{ieeeconf}  % Comment this line out if you need a4paper

\IEEEoverridecommandlockouts                              % This command is only needed if 
\usepackage{graphics} % for pdf, bitmapped graphics files
\usepackage{graphicx} % for pdf, bitmapped graphics files
\usepackage{epsfig} % for postscript graphics files
\usepackage{mathptmx} % assumes a new font selection scheme installed
\usepackage{times} % assumes a new font selection scheme installed
\usepackage{amsmath} % assumes amsmath package installed
\usepackage{amssymb}  % assumes amsmath package installed
\usepackage{comment}
\usepackage{url}

\title{\LARGE \bf
Online identification of skidding modes with interactive multiple model estimation*
}

\author{Ameya Salvi$^{1}$, Pardha Sai Krishna Ala $^{1}$, Jonathon M. Smereka$^{2}$, Mark Brudnak$^{2}$, David Gorsich$^{2}$,\\
Matthias Schmid$^{1}$, Venkat Krovi$^{1}$% <-this % stops a space
\thanks{*This work was supported by the Virtual Prototyping of Autonomy Enabled Ground Systems (VIPR-GS), a US Army Center of Excellence for modeling and simulation of ground vehicles, under Cooperative Agreement W56HZV-21-2-0001 with the US Army DEVCOM Ground Vehicle Systems Center (GVSC).}%
\thanks{**DISTRIBUTION STATEMENT A. Approved for public release: distribution unlimited. OPSEC9037}% <-this % stops a space
\thanks{$^{1}$~All authors are with the Department of Automotive
Engineering at the Clemson University International Center
for Automotive Research (CU-ICAR), Greenville, SC 20607.\{asalvi, pala,schmidm,vkrovi\}@clemson.edu}%
\thanks{$^{2}$~All authors are with the Ground Vehicle Systems Center (GVSC), MI 48397.\{jonathon.m.smereka, mark.j.brudnak,david.j.gorsich\}.civ@army.mil}%
}

\begin{document}

\maketitle
\thispagestyle{empty}
\pagestyle{empty}

%%%%%%%%%%%%%%%%%%%%%%%%%%%%%%%%%%%%%%%%%%%%%%%%%%%%%%%%%%%%%%%%%%%%%%%%%%%%%%%%
\begin{abstract}

Skid-steered wheel mobile robots (SSWMRs) operate in a variety of outdoor environments exhibiting motion behaviors dominated by the effects of complex wheel-ground interactions. Characterizing these interactions is crucial both from the immediate robot autonomy perspective (for motion prediction and control) as well as a long-term predictive maintenance and diagnostics perspective. An ideal solution entails capturing precise state measurements for decisions and controls, which is considerably difficult, especially in increasingly unstructured outdoor regimes of operations for these robots. In this milieu, a framework to identify pre-determined discrete modes of operation can considerably simplify the motion model identification process. To this end, we propose an interactive multiple model (IMM) based filtering framework to probabilistically identify predefined robot operation modes that could arise due to traversal in different terrains or loss of wheel traction.

\end{abstract}

%%%%%%%%%%%%%%%%%%%%%%%%%%%%%%%%%%%%%%%%%%%%%%%%%%%%%%%%%%%%%%%%%%%%%%%%%%%%%%%%
\section{INTRODUCTION}~\label{sec:intro}
~
Skid-steered wheel mobile robots (SSWMRs) are in broad application arenas, ranging from agriculture, mining, construction and forestry. The popularity stems from the necessity of these operating conditions to be facilitated with robots that are rugged (resilient to physical impacts), highly maneuverable (to navigate tight spaces), have significant load-bearing capacity, and are responsive to erratic traction requirements (owing to chances of operating in or getting stuck in soft soils). The lack of steering linkages with an optional addition of independent high-torque hub motors allows SSWMRs to scale up remarkably and thus be an ideal choice for such extreme operations. Although the lack of steering mechanisms simplifies the robot design, it compels it to \textit{skid} its wheels to execute a successful maneuver. Such skidding, while improving the robot's maneuverability by allowing for zero radius turns, makes the robot's motion significantly unpredictable due to the underlying dependence on terrain friction. As a consequence, over the last several years, (partially-) autonomous SSWMRs have therefore functioned largely under a human teleoperator's guidance, coupled with well-instrumented exteroceptive pose-measurements, or by constraining the robot's operation domain to low speeds, small scales, or, flat terrains, where the skidding could be more predictable and the motion thus controllable. Constructing a case for complete autonomy at high speeds and scales therefore entails solving this challenge of characterizing and ultimately controlling skidding for reliable real-time operations.

\begin{figure}
    \centering
    \includegraphics[width=0.95\linewidth]{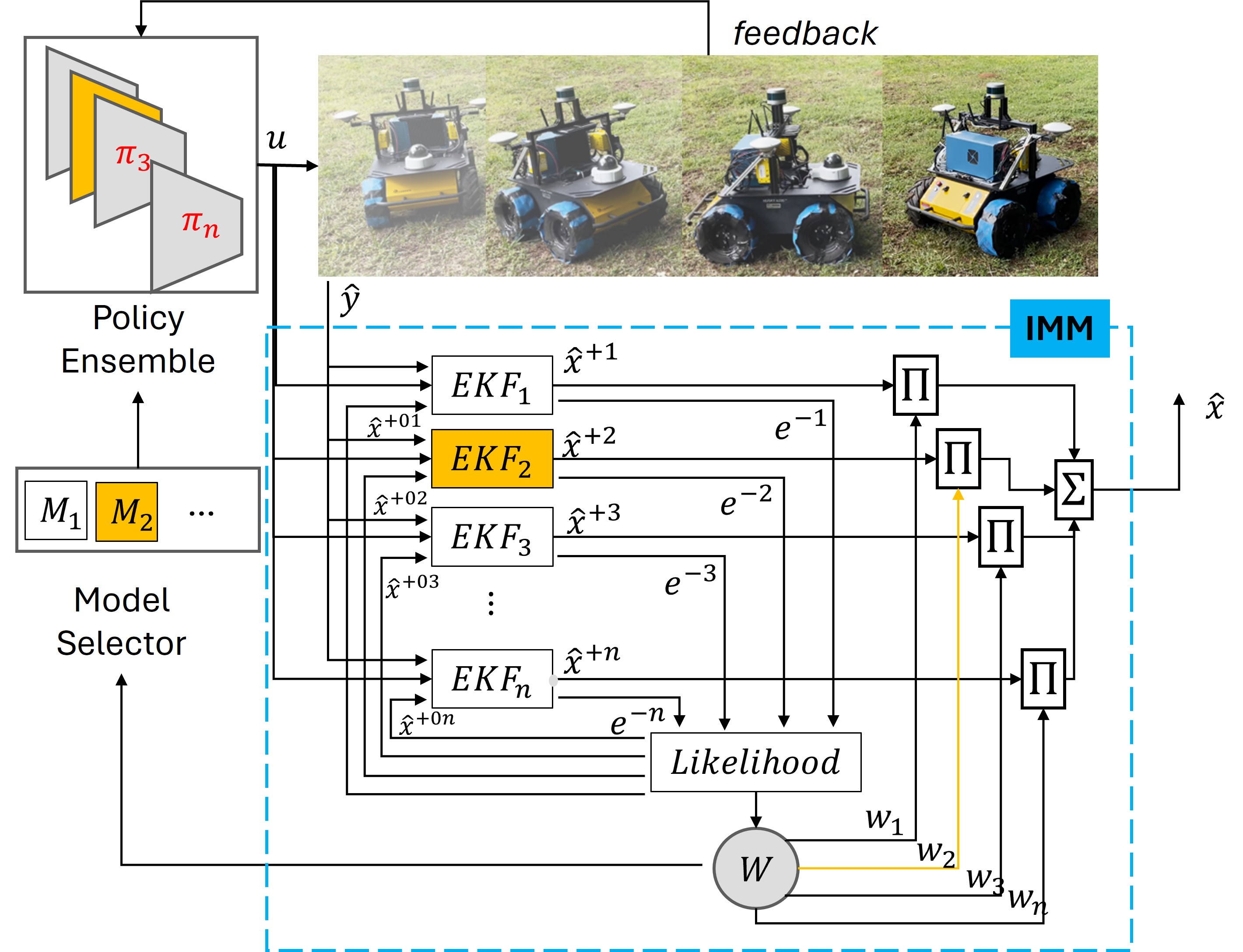}
    \caption{An interactive multiple model estimation framework implemented by employing a bank of parallel extended Kalman filters (EKFs) for identifying motion modes that can be utilized for strategic control policy selection. The highlighted EKF signify the probabilistically identified nearest motion mode that leads to subsequent model selection.}
    \label{fig:IMM-overview}
\end{figure}

The challenges associated with the inability or difficulty to model highly non-linear effects related to friction and terramechanics can be alleviated to an extent with data fitting methods, such as linear and non-linear model fitting, Gaussian processes(GPs) and artificial neural networks (ANNs). However, there are few challenges that need to be dealt with: (i) rapid fluctuation of the environment parameters; and (ii) generalizability of parametric terrain-models to capture the fluctuation; and (iii) rapid sensor-based online adaptation of the model to the new operational  regimes. These challenges are further exacerbated in the case of the robot's mechanical wear that results in significant parametric changes in the physical robot (changes in mass, inertial, tire friction loss). Such changes could be considered minimal (and hence neglected) for structured on-road driving regimes but can significantly impact the performance of SSWMRs operating in unstructured off-road conditions. The real-time identification of such parametric changes in the SSWMR model is thus an important problem whose investigation is crucial for improving the real-time robot motion performance from the point of view of both motion prediction and subsequent decision-making and controls. To this end, the problem of capturing traction loss at SSWMR wheels is investigated in this work. 

In this work, an interactive multiple-model (IMM) based estimation framework is proposed for estimating loss of wheel traction in SSWMRs (as illustrated in figure~\ref{fig:IMM-overview}). IMMs can be categorized as an adaptive estimation technique widely popular in the aerospace community for tracking applications, whose utility as a model identification method in the context of wheel mobile robots (WMRs) is yet to be investigated. The framework introduces a bank of extended Kalman filters (EKFs) that utilize real-time sensor measurements to identify a dominant traction mode of the robot. The dominant model can then be leveraged for improved state prediction or choosing a control policy from an ensemble of control policies, each tuned for a specific dominant traction mode. 

\section{RELATED WORK}~\label{sec:literature}
Over the years, improving motion performance for SSWMRs has been investigated by scholarly contributions in developing robot motion models (to improve model-based controls) and improving robot state estimation (to supplement feedback control methods). Studies have investigated the application of reduced ordered kinematic formulations~\cite{Mandow, Wang, Raibee, Anousaki,Ordonez,Salvi}, both as linear and non-linear approximations for the true robot motion mechanics, while some investigating the validity of these models for extreme conditions~\cite{Baril}. All studies acknowledge the unpredictable nature of skidding and incorporate some form of data fitting approaches to tune their proposed models, introducing the issues associated with quality data collection, processing and accurate fitting. Such models typically face challenges for applications beyond the operation regimes from which data was collected and can be inadequate in accommodating parametric model changes.Concurrently, researchers have investigated dynamic models for SSWMRs that incorporate effects of terrain interactions, ground elevations and intricate tire modeling for formulating dynamic models for real-time utility~\cite{Kozlowski,Huskic,Aguilera,Yu}. While such formulations are more generalizable, the reliability of those for large-scale vehicles is yet to be validated as studies have focused on simulation and analysis or, at the most, on small-scale vehicles in controlled environments.  

Feedback control methods can relax the requirement for accurate models by incorporation of real-time measurements to modulate the control actions. One of the key factors in the performance of feedback control methods is the reliability of the robot's state measurements, which can be impacted by sensor noise. State estimation methods such as Kalman filters and extended Kalman filters allow us to mitigate this issue by recursively updating the state estimate based on real-time measurements and an internal motion prediction model. This functionality comes in handy for SSWMRs, which lack a generalizable motion model.~\cite{Yi-IMU,Yi-Adaptive,Chen,Ordonez} have now explored utilizing state estimation methods for real-time estimation of robot states such as velocities and positions. ~\cite{Ordonez} tries to formulate the problem of such online slip estimation by trying to fit polynomial basis functions in a real-time setting. While closely related to this work, the study focus much more on small model errors where the robot is not subjected to large traction changes. It is noteworthy to mention work in~\cite{McKinnon} which conversely investigates on-line learning for large model changes such as robot mass and inertia where robot skidding is not a primary focus of the study. 

Significant changes in motion models (similar to traction losses for SSWRMs) have been investigated in other engineering application arenas for fault detection and identification (FDI) and fault tolerant control (FTC). These applications rely on pre-identified failure modes and utilize adaptive estimation methods for identifying those in real-time and subsequently switching the control strategies~\cite{Raman, Gill, Tudoroiu}. Systematic implementation and validation of these methods as traction mode identification strategy for SSWMRs is a promising research area that can significantly improve the safety and reliability of SSWMR motion.

\begin{figure}
\vspace{0.5cm}
    \centering
    \includegraphics[width=0.95\linewidth]{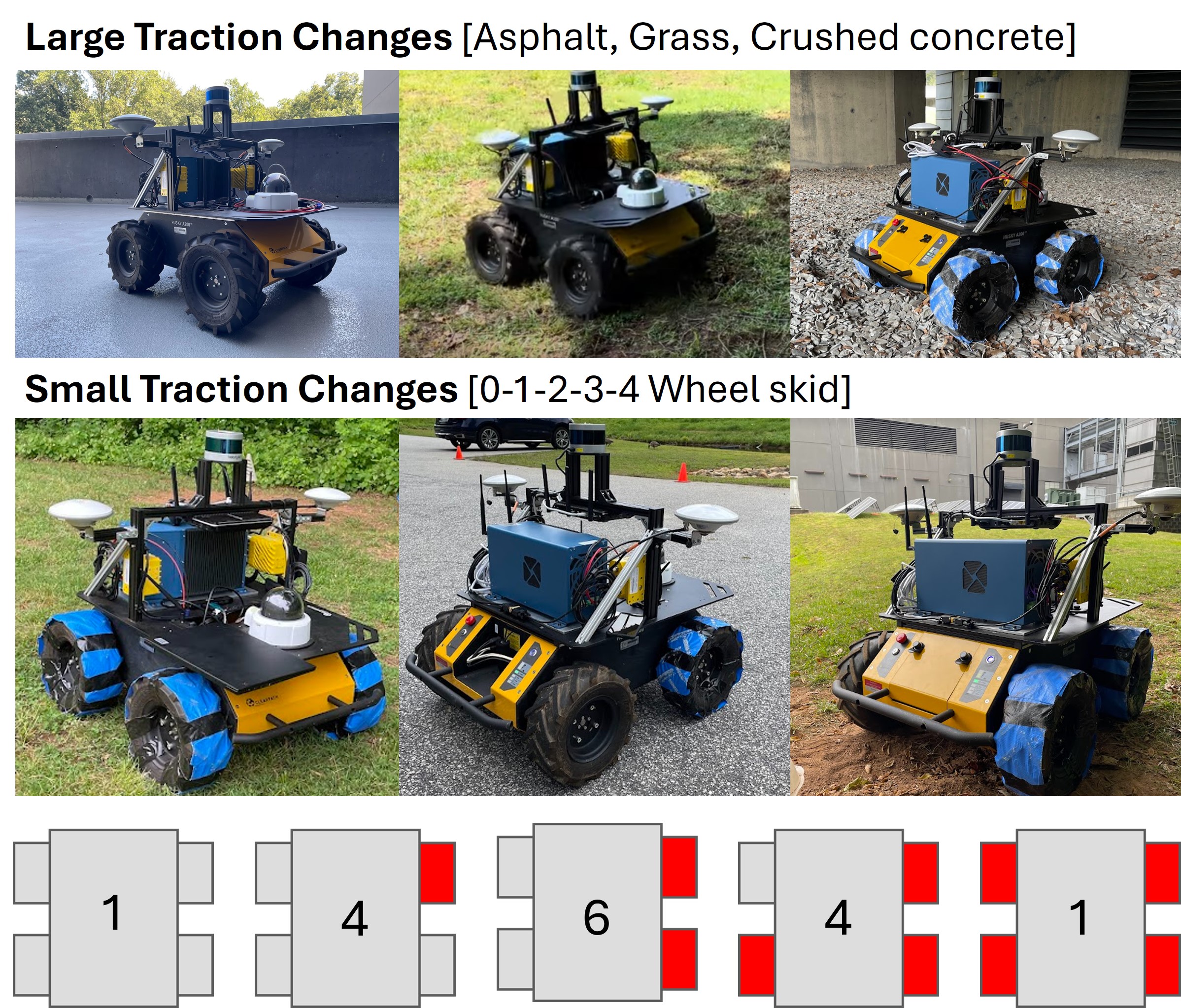}
    \caption{Numerous deterministic motion modes realized due to changes in available traction. (Top) Modes for large traction changes (LTC) for operation in asphalt, grass and crushed concrete. (Bottom) Modes for small traction changes (STC) arising due to loss of traction at one or several wheels. Traction losses are experimentally induced by wrapping wheels with polyethylene tarps resulting in the following combinations : 0-1-2-3-4 wheels skidding (total 16 combinations) with \textit{red} wheels indicating covering with tarp.}
    \label{fig:traction-modes}
\end{figure}

Figure~\ref{fig:traction-modes} illustrates the predefined traction modes identified in this work. \textit{Large traction changes} are considered when the robot traverses different surfaces and consequently showcases significant variations in the motion dynamics. These variations are distinct enough to be captured even with low-fidelity sensor measurements such as the inertial measurement units (IMUs). \textit{Small traction changes} are considered when one or multiple robot wheels wear out while traversing the same surface. This work has constructed such a scenario by covering the wheels with a polyethylene tarp, thus reducing the available traction. Small traction changes can potentially be difficult to capture solely with IMU data due to measurement noise and may require stable measurements from sensors such as global positioning systems (GPS).

%Single or multiple wheels are covered with polyethylene tarps, inducing friction loss on loose soil compared to actual wheels. A total $16$ such pre-defined combinations can be realized for which a $1$ to $16$ operation modes can be defined depending requested control resolution. For all processes, inertial measurement units (IMUs) and wheel encoders have been identified at sensors of choice to allow the incorporation of this framework within dead-reckoning applications.

\section{MOTION MODELS}~\label{sec:MotionModels}
~
Within the estimation framework, motion models allow the propagation of the state and error estimates, which are later updated with the sensor measurements. Thus, the choice of motion models can be dependent on (a) the accuracy of state estimates required, (b) computational complexity, and (c) sensors and measurements available, amongst many other things. Furthermore, the ability to capture the variations from the sensor measurements due to changes in operation mode depends on the sensor's resolution. Thus certain variations could easily be captured with lower-quality sensors such as the IMU whereas others may require more precision instruments such as the GPS. To this end, two distinct frameworks, one for large traction changes (LTC) based on IMU data and another for small traction changes (STC) based on GPS data, have been proposed in this work.

The two frameworks adopt the standard structure for recursive state estimation in discrete time domain with the difference in choice of states and thus their respective motion models.

\begin{equation}
\label{eq:ModelStruct}
    \begin{aligned}
    &\mathbf{x}_{k+1} = \mathbf{x}_{k} + \mathbf{f}_{i}(\mathbf{x}_{k},\mathbf{q}_{k})\Delta t + \mathbf{w}_{k} \\
    &\Tilde{\mathbf{y}}_{k+1} = \mathbf{h}(\mathbf{x}_{k}) + \mathbf{v}_{k}
    \\
    &\text{where,} \\
    &\mathbf{x} = [V_B~\omega_B]^{T}~(LTC)~\text{OR}~\mathbf{x} = [X_F~Y_F~\theta_F]^{T}~(STC) \\
     \end{aligned}
\end{equation}

Equation~\ref{eq:ModelStruct} illustrates the propagation for state $\mathbf{x}\in\mathbf{R}^{N}$ for any given sampling step, $k$, discretized at the time interval $t$. The size of the state vector, $N$, depends on the choice of framework which could be $2$ for estimating robot longitudinal velocity ($V$) and angular velocity ($\omega$) expressed in the body frame $B$, or, could be $3$ for estimating robot's position and orientation ($X,Y~\text{and}~\theta$) expressed in the space frame $F$. The choice of the states depends on the ease of availability of robot sensor measurements which are utilized from the IMU measurements (typically measured in body frame) and GPS measurements (typically available in space frame). $w \in\textbf{R}^{N \times N} ~\text{and}~v \in\textbf{R}^{N \times N}$  are process and measurement noise covariances. $h~\in\mathbf{R}^{N \times N}$ is the measurement model represented as an identity matrix in this work. $\mathbf{f}$ is state transition function or the robot motion model identified with a combination of analytical and data fitting methods to represent $i$ distinct modes of operation. The data used for the fitting process comes from a series of motion maneuvers such as skidpad, clothoids and sinusoidal, conducted a priori on the different robot configurations.

%are the robot states to be estimated, $\mathbf{u}$ is the robot control input, $\mathbf{f}(\mathbf{x}(t),\mathbf{u}(t))$ is the state evolution model, $\mathbf{h}(\mathbf{x}(t))$ is the measurement model, and, $\mathbf{w}~\text{and}~\mathbf{v}$ are the process and measurement noises. The measurement model is chosen as an identity matrix, signifying the direct availability of measurement data. The explicit identification of $\mathbf{f}$ is done following a combination of analytical and data-fitting approaches adopted in the literature. The data used for fitting comes from a sequence of clockwise and counter-clockwise clothoids (constant curvature curves) as motion maneuvers in the various operation modes. Thus, a data bank containing $19~\times~2$ maneuvers for $19$ models is collected. 

\subsection{Models for Large Traction Changes (LTC)}~\label{subsec:LTCModels}

%Significant changes in traction can noticeably impact robot dynamics, particularly during high-speed maneuvers or on diverse surfaces. These changes can be reliably captured using IMU data, allowing for the estimation of the states $\mathbf{x} = [V~\omega]^T$, where $V$ and $\omega$ represent the robot’s longitudinal velocity (in $\text{m/s}$) and angular velocity (in $\text{rad/s}$), respectively. 
%Although IMU accelerometers measure acceleration (in $\text{m/s}^2$), it is straightforward to compute velocity through integration. This allows for accurate real-time tracking of large traction changes as they manifest in velocity and angular rate deviations.

Identifying models for large traction changes requires a state evolution function for the states $V~(m/s)$ and $\omega~(rad/s)$ influenced by the robot control inputs, $\dot{\phi}_{L}~(rad/s)~\text{and}~\dot{\phi}_{R}~(rad/s)$, which are the left and right wheel velocities. This has been achieved by fitting streams of data collected from all the maneuvers in the LTC configurations for coming up with linear models $A_{2 \times 2}$ and $B_{2 \times 2}$.

\begin{equation}
\label{eq:ModelStructLTC}
\begin{aligned}
    &\begin{bmatrix}
        \dot{V} \\
        \dot{\omega}
    \end{bmatrix}
    = A_{i}
    \begin{bmatrix}
        V \\
        \omega
    \end{bmatrix}
    + B_{i}
    \begin{bmatrix}
        \dot{\phi}_{L} \\
        \dot{\phi}_{R}
    \end{bmatrix}\\[1 ex]
    &i={\text{Asphalt, Grass, Crushed concrete}}
%    e &= (A - A_i)\mathbf{x} + (B - B_i)\mathbf{u}, \\[1ex]
%    \mathbf{w}_i &\sim \mathcal{N}(\mu_i, \sigma^2_i)
\end{aligned}
\end{equation}

It is noteworthy to mention that another fitting approach involving fitting only one linear matrix $A$ (and $B$) from all the data, but with different noise covariances matrices $\mathbf{w}_{i}$ (from the propagation equation~\ref{eq:ModelStruct}) was also explored and provided similar results.

%where $A$ and $B$ are the global model matrices, and $A_i$ and $B_i$ are local models fitted for specific traction conditions (such as asphalt, grass, or crushed concrete). The error $e$ between the global and local models is treated as process noise, modeled as a normal distribution $\mathcal{N}_i(\mu_i, \sigma^2_i)$. 

%The global model is fitted using linear least squares on the entire dataset, while local models $A_i$ are trained specifically for each surface. This approach enables the detection of significant changes in traction, adapting the model to account for surface-specific dynamics.

\subsection{Models for Small Traction Changes (STC)}~\label{subsec:STCModels}

While significant traction changes are more easily detected using IMU data, identifying small traction changes presents a greater challenge due to the low signal-to-noise ratio, especially at lower speeds. Despite their subtlety, small changes can accumulate over time, leading to increased wear on the robot and unnecessary energy consumption. For small traction changes, the state vector is modified to track the robot's inertial pose: $\mathbf{x} = [X~Y~\theta]^T$, where $X~(m)$ and $Y~(m)$ are the robot’s position, and $\theta~(rad)$ is the heading angle. 

The standard pose estimation model used for robot localization has been adopted for this approach~\cite{SiegwartAMR,ThrunPR}. The slight modification involves introduction of tuning constants $k$ and $m$ in the non-linear equations relating the body velocity to the inertial velocity. The different tuning constants signify the $i = 16$ different physical configurations for wheel slips as illustrated in fig.~\ref{fig:traction-modes}. 

\begin{equation}
\label{eq:ModelStructSTC}
\left.\begin{aligned}
    \dot{X} &= k_i \cdot \cos(\theta) \cdot u_1, \\
    \dot{Y} &= k_i \cdot \sin(\theta) \cdot u_1, \\
    \dot{\theta} &= m_i \cdot u_2
    \end{aligned}\right\}
    \begin{aligned}
    u_1 =& \frac{r (\dot{\phi}_L + \dot{\phi}_R)}{2} \\
    u_2 =& \frac{r (-\dot{\phi}_L + \dot{\phi}_R)}{b} 
    \end{aligned}
\end{equation}

where $u_1$ is the requested longitudinal body velocity and $u_2$ is the requested angular velocity. Here, $r$ represents the wheel radius, and $b$ is the robot’s track width. The non-linear representation in equation~\ref{eq:ModelStructSTC} can be partially differentiated with the robot states $X,Y,\theta$ and the inputs $u_{1}~\text{and}~u_{2}$ for a Jacobian representation conforming with equation~\ref{eq:ModelStruct}.

\begin{comment}

These conditions are categorized into four representative slip modes:
\begin{enumerate}
    \item \textbf{Baseline}: No slip (all wheels maintain full traction).
    \item \textbf{Two Right Wheels Slipping}: Reduced traction on both of the right wheels.
    \item \textbf{Two Front Wheels Slipping}: Reduced traction on both of the front wheels.
    \item \textbf{All Four Wheels Slipping}: All wheels experience slipping.
\end{enumerate}

\end{comment}

%The parameters $k_i$ and $m_i$ capture the degradation in body velocity $V$ and angular velocity $\omega$, respectively, for each slip mode. This allows the model to dynamically adjust the state estimation and control inputs based on observed slip, ensuring more accurate predictions of the robot's motion over time. The non-linear representation in equation~\ref{eq:ModelStructSTC} can be partially differentiated with the robot states $X,Y,\theta$ and the inputs $u_{1}~\text{and}~u_{2}$ for a Jacobian representation conforming with equation~\ref{eq:ModelStruct}, which is a standard implementation of the robot localization model as showed in~\cite{SiegwartAMR,ThrunPR}.

\section{INTERACTIVE MULTIPLE MODEL ESTIMATION}~\label{sec:IMMM}

%As discussed, the motion of SSWMRs across different terrain conditions requires using different motion models to capture system dynamics accurately. 
In the multiple model (MM) estimation approaches, it is assumed that the model describing the system dynamics emergees from a predefined set of models. The static MM estimator does not permit switching between different models among these approaches. In contrast, dynamic MM estimators, such as the Interactive Multiple Model (IMM) filters, enable real-time transitions between different system models \cite{Bar Shalom}. IMM framework allows SSWMRs to adapt to changing terrain conditions, failure modes of wheels, and several other influencing factors by dynamic switching between different models. Despite its suboptimal nature, the IMM filter efficiently provides state estimates and active model identification, making it suitable for real-time applications. Based on the available models, the IMM filter employs a parallel bank of filters, each corresponding to a specific system model.

Considering a bank of N filters, the IMM filtering algorithm for $k^{th}$ filtering step can be classified into the following stages: \newline
\begin{enumerate}
    
 \item \textbf{Interaction/mixing}: During this phase, the input to the model matched filter is obtained by weighted mixing of updated states of all filters. The mixing probability, $\mu_{i|j}$, is defined as the probability of model $M_{i}$ being in effect at $k-1$ given that model $M_j$ is in effect at $k$ and conditioned on measurements until $k-1$.

\begin{equation}
    \label{eq:MixingProbability}
    \begin{aligned}
        &\mu_{k-1}^{i|j} = \frac{p_{ij}\mu_{k-1}^{i}}{c_{j}}  \hspace{0.8cm}     i, j \in [1, ..., N]\\
    \end{aligned}
\end{equation}

$p_{ij}$ defines the transitional probability from $M_{i}$ to $M_{j}$. The probability of \textit{$i^{th}$} model , $M_{i}$ being in effect at $k-1$ is given as $\mu_{i}(k-1)$. ${c_{j}}$ is the normalizing factor
\begin{equation}
    \label{eq: Mixing probability normalizing factor}
    \begin{aligned}
        & c_{j} =  \sum_{i = 1}^{i = N} p_{ij}\mu_{k-1}^{i} \\
    \end{aligned}
\end{equation}
%Compared to the second-order generalized pseudo-Bayesian approach (GPB2), IMM allows for interaction between filters at the beginning of each cycle and provides input to each filter based on the mixing probabilities.

 \item \textbf{Mixed states and covariances}: Now that the mixing probabilities are obtained, the next step in the cycle is to obtain the input to the model-matched filter $j$ in terms of state estimate and state error covariance.
\begin{equation}
    \label{eq:MixedState}
    \begin{aligned}
        & \hat{x}_{k-1}^{+0j} =  \sum_{i = 1}^{ N} \hat{x}_{k-1}^{+i} \mu_{k-1}^{i|j} \hspace{1.4cm}  j = 1, ..., N
    \end{aligned}
\end{equation}
Similarly, for all the filters, mixed covariance is calculated as the following
\begin{equation}
    \label{eq:MixedCov}
    \begin{aligned}
        & P_{k-1}^{+0j} =  \sum_{i = 1}^{N}  \mu_{k-1}^{i|j} \left[P_{k-1}^{+i} + [\hat{x}_{k-1}^{+i}-\hat{x}_{k-1}^{+0j}][\hat{x}_{k-1}^{+i}-\hat{x}_{k-1}^{+0j}]' \right]
    \end{aligned}
\end{equation}

 \item \textbf{Model-matched filtering}: Based on the matched model, mixed states and covariances, depending on the system models, state estimation filters such as the Kalman filter, and extended Kalman filter (EKF) are used for state estimation. The updated state estimates $\hat{x}_{k}^{+j}$ and error covariances $P_{k}^{+j}$ are obtained  for the model matched filter, $M_{j}$. The likelihood function $\Delta_{k}^{j}$, which defines the likeliness of the measurements based on the matched model, is calculated.
\begin{equation}
    \label{eq:Likelihood}
    \begin{aligned}
   %     & \Delta_{k}^{j} =  \frac{1}{(det(2\pi E_{k}^{-j}))^{0.5}} e^{-\frac{(\Tilde{y}_k-H\hat{x}_{k}^{-j})'(E_{k}^{-j})^{-1}(\Tilde{y}_k-%H\hat{x}_{k}^{-j})}{2}}
        & \Delta_{k}^{j} =  \frac{\exp\left(-\frac{(\tilde{y}_k - H\hat{x}_{k}^{-j})'(E_{k}^{-j})^{-1}(\tilde{y}_k - H\hat{x}_{k}^{-j})}{2}\right)}{\sqrt{\det(2\pi E_{k}^{-j})}} 
    \end{aligned}
\end{equation}

\item \textbf{Model probability update}: The model probability is the posterior probability of model $j$, $M_{j}$ conditioned on the entire measurement data set up to $k$. The likelihood function of the model matched filter is used to update the model probability.
\begin{equation}
    \label{eq:ModalProbability}
    \begin{aligned}
        &\mu_{k}^{j} = \frac{\Delta_{k}^{j} \mu_{k-1}^{j}}{\sum_{i=1}^{N} \Delta_{k}^{i} \mu_{k-1}^{i}} \\
    \end{aligned}
\end{equation}

\item \textbf{Combined state estimates and covariances}: The weighted sum of state estimates and covariances is calculated based on the model probabilities and each filter's updated state estimates and error covariances. 
\begin{equation}
    \label{eq:CombinedState}
    \begin{aligned}
        & \hat{x}_{k}^{+} =  \sum_{i = 1}^{N} \hat{x}_{k}^{+i} \mu_{k}^{i} \hspace{1.4cm}  j = 1, ..., N
    \end{aligned}
\end{equation}
\begin{equation}
    \label{eq:CombinedCov}
    \begin{aligned}
        & P_{k}^{+} =  \sum_{i = 1}^{N}  \mu_{k}^{i} \left[P_{k}^{+i} + [\hat{x}_{k}^{+}-\hat{x}_{k}^{+i}][\hat{x}_{k}^{+}-\hat{x}_{k}^{+i}]' \right]
    \end{aligned}
\end{equation}

\end{enumerate}

The model probabilities or \textit{weights} calculated in step $4$ identify the dominant operational mode of the robot allowing to detect the change in the robot's physical configuration. The dynamic switching further allows to capture the continuing physical changes whose performance has been investigated in the next section.

\section{RESULTS}~\label{sec:Results}

The results section covers the two sets of experiments for (A) varying terrains (LTC), and (B) slip at various wheels (STC). Each experiment is further investigated for identification for a singular (non-varying robot configuration) and switching mode. The singular mode experiment investigates the prolonged stability of the identification framework where as the switching mode investigate the responsiveness of the identification framework to model changes. All evaluations are done on an outward spiral maneuver in which the robot turning radius increases at a constant rate. 

%While  can be identified in a real-time deployment, testing for continuously varying configurations is difficult as manual intervention is necessary to physically modify the robot configuration. To this end, the evaluation dataset has been constructed by concatenating together measurements (recorded \textit{rosbags}) captured in each mode. These measurements are then played in real-time and the IMM algorithm tries to estimate as if the robot configuration switches automatically.

\subsection{Mode identification for varying terrain (LTC)}~\label{subsec:ResultsLTC}

\begin{figure*}
\vspace{0.5cm}
    \centering
    \includegraphics[width=0.95\linewidth]{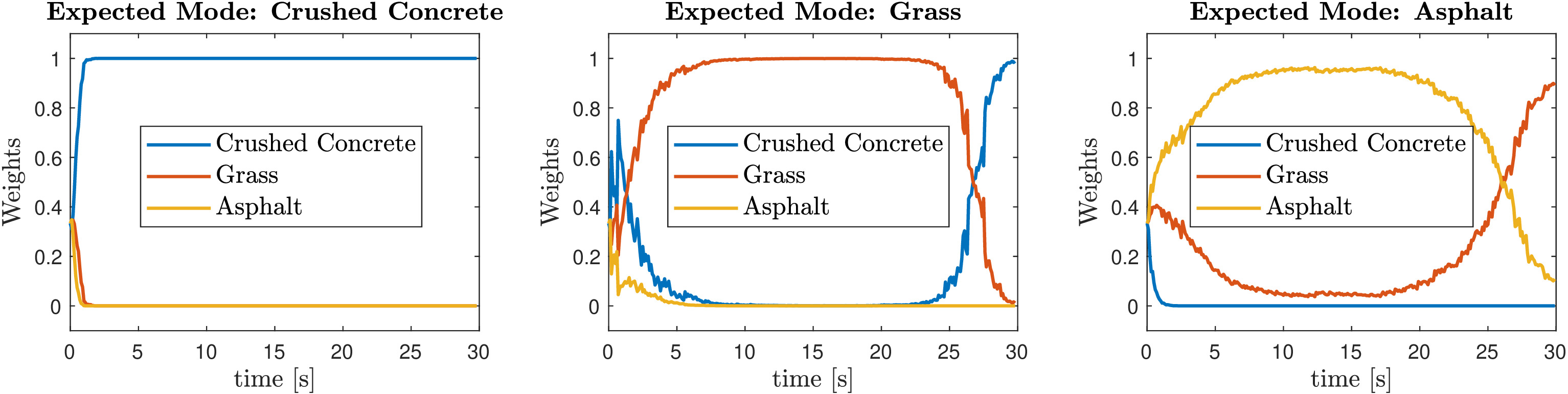}
    \caption{Singular mode identification on three sets of terrain where the algorithm is expected to stay stable in one of the operation modes. (a) Expected mode: Crushed concrete, (b) Expected mode: Grass, (c) Expected mode: Asphalt.}
    \label{fig:ModelWeightsLTC}
\end{figure*}

Model identification on varying terrains is investigated by utilizing the motion models illustrated in eq.~\ref{eq:ModelStructLTC} and the robot velocity measurements $V_{b}~\text{and}~\omega_{B}$ with a Kalman filter implementation within the IMM framework. Three terrain variations, including asphalt, grass and crushed concrete are investigated for this set of experiments. Figure~\ref{fig:ModelWeightsLTC} illustrates the framework's performance for identifying the model configurations within a singular mode. 

Figure~\ref{fig:ModelWeightsLTC} illustrates that the framework promptly identifies the crushed concreted operation mode (in less than $5$ seconds) and stays stable in the same mode as compared to the other two modes. For grass and asphalt, the framework takes much longer to identify the operation mode and also struggles to stay stable in the same. It can be seen from the second and third subplots of fig.~\ref{fig:ModelWeightsLTC} that the framework faces challenges to distinguish between grass and crushed concrete, and, between grass and asphalt, and confuses for the incorrect operation mode towards the end of the cycle. One potential reason for the confusion towards the end of the cycle could be because of the nature of the maneuver in which the robot travels \textit{almost} straight towards the end. While the robot dynamics are significantly different on the sharp curves, the algorithm may face challenges on straights where the robot dynamics can be indistinguishable between operation modes.

%This is likely due to the similarities between asphalt and grassy surfaces in terms of their solid, smooth, and stable nature. In contrast, the crushed concrete surface, being loose and rough, offers better traction due to its rugged, coarse texture, particularly in dry conditions.
%illustrates that the IMM filter identifies the model associated with the crushed concrete surface less than 5 seconds of receiving measured data. Compared to other surfaces, crushed concrete's distinct characteristics allow the IMM filter to recognize the model promptly. For the grassy surface, the IMM filter takes slightly longer to identify the corresponding model, as indicated in the second subplot of  Figure~\ref{fig:ModelWeightsLTC}. At both the start and end of the cycle, the model associated with the crushed concrete surface is identified, likely due to the WMR’s reduced traction and inertia.  The asphalt surface is immediately distinguished from the other surfaces in the third plot. However, toward the end of the cycle, the model corresponding to the grassy surface takes precedence over the actual asphalt surface. 

\begin{figure}
    \centering
    \includegraphics[width=0.95\linewidth]{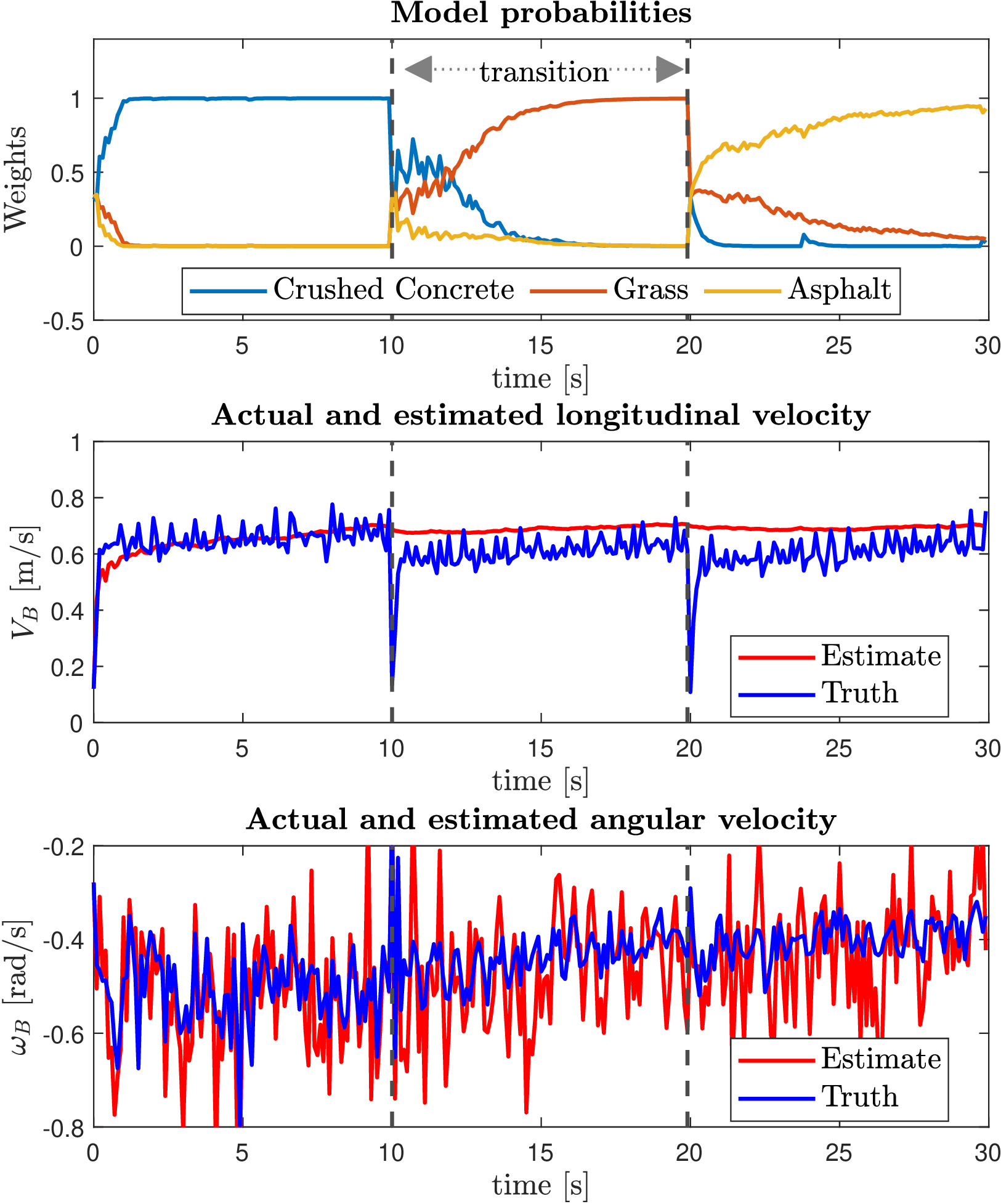}
    \caption{Varying mode identification on three sets of terrain where algorithm is expected to switch real-time within the operation modes. Vertical dashed lines at $t = 10s$ and $t = 20s$ signify physical change in the operation mode (Crushed concrete - to Grass - to Asphalt).}
    \label{fig:ModelWeightsLTC-Switch}
\end{figure}

Figure~\ref{fig:ModelWeightsLTC-Switch} illustrates the successful switching between the operation modes along with the state estimates ($V_{B}~\text{and}~\omega_{B}$) for a composed maneuver of $30$s with $10$s in each mode. The results indicated performance similar to singular mode where the framework identifies crushed concrete fairly quickly but requires a while to identify the other two modes. It can also be observed that the state estimates are much better in the case of crushed concrete, most likely due to it significant distinction from the other two modes (coarse and granular as compared to much tightly pack texture for grass and asphalt).

%performance for continuously varying operation modes. Here, the filter's longitudinal and angular velocity estimates are shown for the different terrain conditions. When the WMR traverses the crushed concrete surface, it can travel with relatively higher velocity due to the rugged, coarse nature of the terrain. IMM filter tracks the states with relative accuracy for all the different terrain conditions. The covariance of the process noise for the velocity model is lower than the angular velocity model, and the velocity estimate performs significantly better in tracking the states compared to the angular velocity estimates. The angular velocity error fluctuates around the zero mean within its $3\sigma$ limits.  

\subsection{Mode identification for varying skid (STC)}~\label{subsec:ResultsSTC}

\begin{figure*}
\vspace{0.5cm}
    \centering
    \includegraphics[width=0.95\linewidth]{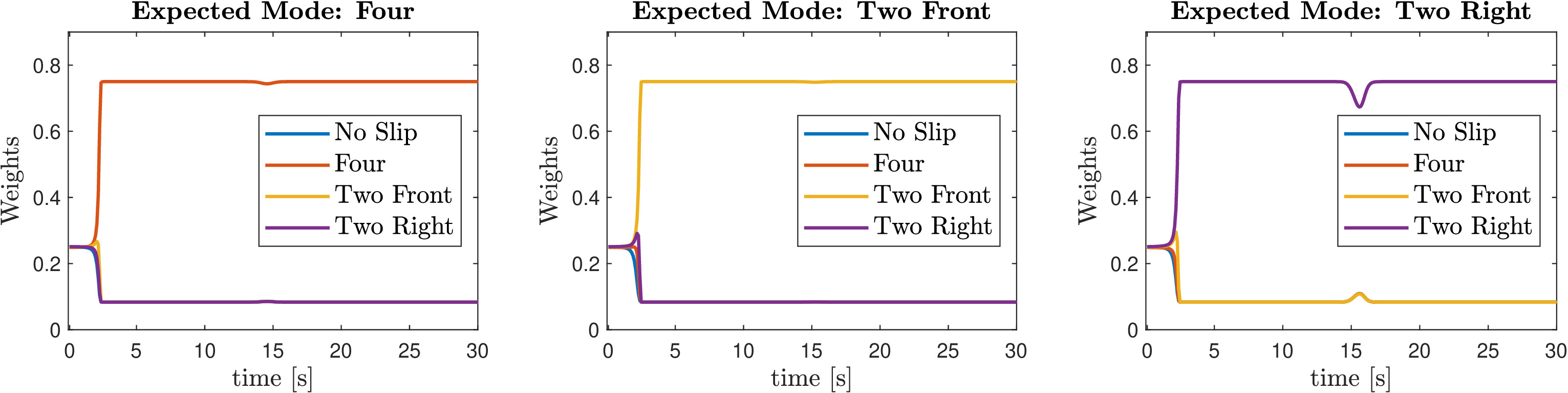}
    \caption{Singular mode identification on four sets of wheel slip modes where algorithm is expected to stay stable in one of the operation modes. (a) Expected mode : Four wheel skid, (b) Expected mode :Two Front wheel skid, (c) Expected mode : Two right wheel skid.}
    \label{fig:ModelWeightsSTC}
\end{figure*}

As discussed in the section~\ref{sec:MotionModels}, identifying skidding modes purely on IMU data challenging and thus adopting the robot pose as the estimation states for skidding mode identification. Since the robot pose is expressed as non-linear model as shown in eq.~\ref{eq:ModelStructSTC}, an EKF based implementation of the IMM framework was utilized. Figure~\ref{fig:ModelWeightsSTC} illustrates the performance of the estimation framework for detecting the model weights for singular operation modes. To limit the scope of the investigation, the estimation framework was initialized with a mixture of four wheel slip modes : (a) No wheels slipping, (b) Front two wheels slipping, (c) Right two wheels slipping, and, (d) All wheels slipping.

It was observed that the framework could successfully identify the operation modes with the two right, two front and all four wheels skidding. The estimation framework struggles slightly to differentiate between the two right wheels and two front wheels skidding around $t = 15s$, but stabilizes itself fairly quickly. It should be noted that the framework struggled to identify the baseline model and almost always incorrectly detected the four wheel skidding model. This could potentially be due to the fact that the baseline case (no wheel skidding) and four wheel case (all four wheels skidding) are similar in the terms that all wheels have same friction coefficient in a particular mode. As a consequence, the behavior exhibited can be indistinguishable for the IMM framework and performance improvement may required more data for motion model fitting and considerable tuning of process and measurement noise. 

\begin{figure}
    \centering
    \includegraphics[width=0.95\linewidth]{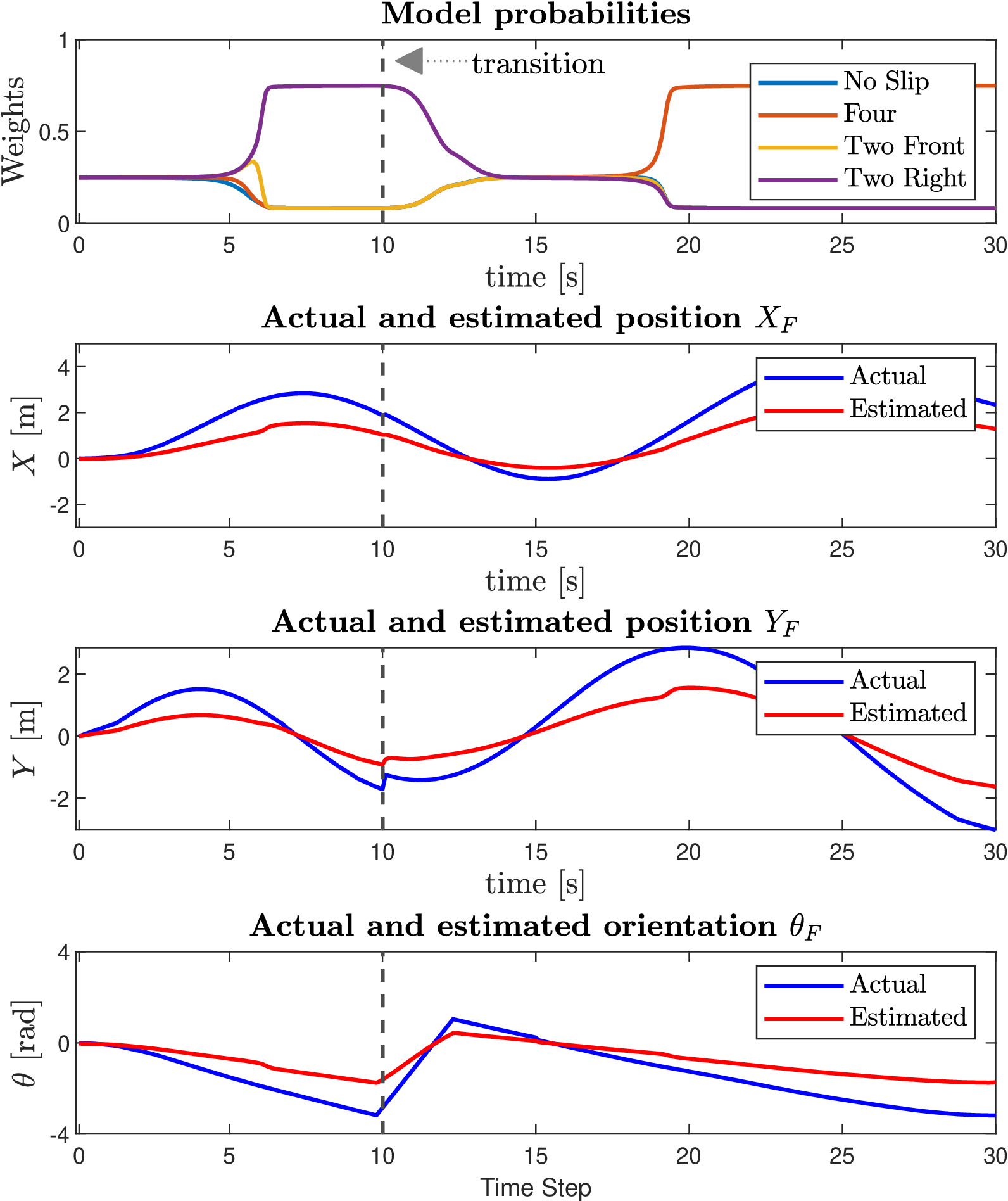}
    \caption{Varying mode identification on two sets of wheel skid configurations where algorithm is expected to switch real-time within the operation modes. Vertical dashed lines at $t = 10s$ signify physical change in the operation mode (Right two wheels skid - to all wheels skid).}
    \label{fig:ModelWeightsSTC-Switch}
\end{figure}

Figure~\ref{fig:ModelWeightsSTC-Switch} illustrates the model performance in the case of continuously switching modes. The two modes selected for this test are first two front wheels skidding and then all four wheels skidding to elucidated partial and complete loss of traction. The true model transition occurs at $t = 10s$ which is depicted by vertical dashed line. It can be seen that the frameworks takes almost 10 seconds after the model transition to detect the change in the mode. This is significantly different as compared to performance of the former framework where model switches happen fairly quickly. The change in the performance could be attributed to the fact that small model changes are significantly difficult to distinguish even with GPS data.

\section{Discussion}~\label{sec:Conclusion}
In this work, an IMM based framework for identifying robot skidding modes has been investigated with utilizing either (a) the robot body velocity , or (b) the robot pose as the underlying states to be estimated. It was observed that skidding mode identification can be considerably depended on the magnitude distinction in both the robot motion models and measurement signals. For slow velocity robots like the Husky, the framework requires substantial tuning and still be sometimes fragile to noisy measurement signals. Improving distinction between the models with different basis for model fitting or varied maneuvers for data collection can potentially be investigated to check for performance improvement. Current plans are to utilize this framework for larger skid-steered robots such as the Clearpath Warthog which operated across a larger range of speeds.

\addtolength{\textheight}{-12cm}   % This command serves to balance the column lengths

%%%%%%%%%%%%%%%%%%%%%%%%%%%%%%%%%%%%%%%%%%%%%%%%%%%%%%%%%%%%%%%%%%%%%%%%%%%%%%%%

%%%%%%%%%%%%%%%%%%%%%%%%%%%%%%%%%%%%%%%%%%%%%%%%%%%%%%%%%%%%%%%%%%%%%%%%%%%%%%%%

%%%%%%%%%%%%%%%%%%%%%%%%%%%%%%%%%%%%%%%%%%%%%%%%%%%%%%%%%%%%%%%%%%%%%%%%%%%%%%%%

\section*{ACKNOWLEDGMENT}

This work was supported by the Virtual Prototyping of Autonomy Enabled Ground Systems (VIPR-GS), a US Army Center of Excellence for modeling and simulation of ground vehicles, under Cooperative Agreement W56HZV-21-2-0001 with the US Army DEVCOM Ground Vehicle Systems Center (GVSC).

\section*{CONTRIBUTIONS}

MS provided the conceptualization and the theoretical
framework for the research presented in this work and supervised
its development. AS and PA modelled the system, performed the coding,
analyses and development and deployment of the framekwork. VK, JS, MB and DG served as advisors and supervised the alignment of the
presented work with the larger context of the underlying VIPR-GS projects
and contributed towards the funding of this research.

%%%%%%%%%%%%%%%%%%%%%%%%%%%%%%%%%%%%%%%%%%%%%%%%%%%%%%%%%%%%%%%%%%%%%%%%%%%%%%%%

\end{document}